\documentclass{IOS-Book-Article}
\usepackage[dvipsnames]{xcolor}
\usepackage{mathptmx}
\usepackage{graphicx}
\usepackage{amsmath}
\usepackage{subcaption}
\usepackage{graphicx}
\usepackage{rotating} 

\usepackage{soul}\setuldepth{article}
\def\hb{\hbox to 10.7 cm{}}

\begin{document}

\pagestyle{headings}
\def\thepage{}

\begin{frontmatter}              

\title{Sentiment and Emotion-aware Multi-criteria Fuzzy Group Decision Making System}

\markboth{}{July 2024\hb}



\author[A]{\fnms{Adilet} \snm{Yerkin} \orcid{0009-0007-6228-7296}},
\author[A]{\fnms{Pakizar} \snm{Shamoi}%
\thanks{Corresponding Author: Pakizar Shamoi; E-mail:
p.shamoi@kbtu.kz} \orcid{0000-0001-9682-0203}},
\author[A]{\fnms{Elnara} \snm{Kadyrgali} \orcid{0009-0000-4732-2806}}

\address[A]{School of Information Technology and Engineering, Kazakh-British Technical University, Almaty, Kazakhstan}

\begin{abstract}
In today’s world, making decisions as a group is common, whether choosing a restaurant or deciding on a holiday destination. Group decision-making (GDM) systems play a crucial role by facilitating consensus among participants with diverse preferences. Discussions are one of the main tools people use to make decisions. When people discuss alternatives, they use natural language to express their opinions. Traditional GDM systems generally require participants to provide explicit opinion values to the system. However, in real-life scenarios, participants often express their opinions through some text (e.g., in comments, social media, messengers, etc.). This paper introduces a sentiment and emotion-aware multi-criteria fuzzy GDM system designed to enhance consensus-reaching effectiveness in group settings. This system incorporates natural language processing to analyze sentiments and emotions expressed in textual data, enabling an understanding of participant opinions besides the explicit numerical preference inputs. Once all the experts have provided their preferences for the alternatives, the individual preferences are aggregated into a single collective preference matrix. This matrix represents the collective expert opinion regarding the other options. Then, sentiments, emotions, and preference scores are inputted into a fuzzy inference system to get the overall score. The proposed system was used for a small decision-making process - choosing the hotel for a vacation by a group of friends. Our findings demonstrate that integrating sentiment and emotion analysis into GDM systems allows everyone’s feelings and opinions to be considered during discussions and significantly improves consensus among participants. 

\end{abstract}

\end{frontmatter}
\markboth{July 2024\hb}{July 2024\hb}

\section{Introduction}
Nowadays, group decision-making (GDM) is a part of many people's everyday lives.  When people discuss making decisions, they use natural language to express their opinions. Traditional GDM systems generally require participants to provide explicit opinion values to the system \cite{morente2018analysing}. However, in real-life scenarios, participants often express their opinions through some text (for example, in comments, social media, messengers, etc.), which will be a challenge to computer systems. Approaches to addressing this challenge include sentiment analysis techniques, fuzzy logic, and fuzzy sets. Using sentiment analysis, we can measure the user's sentiment and subsequently determine the preference value assigned to a particular alternative by the user. This allows the computer system to incorporate the expressed opinions effectively into the decision-making process. Using fuzzy logic and fuzzy sets, groups can make decisions when there is uncertainty or ambiguity. 



Discussions are one of the main tools people use to make decisions. When a decision process involves a group of people and requires multiple points of view to consider, it is commonly conducted in groups of people. GDM \cite{blin1978individual} systems are tools and processes that allow a group of people to work together to reach a consensus acceptable to all group members. GDM systems can be useful where multiple perspectives need to be considered to arrive at a decision acceptable to all group members \cite{Herrera1996}, \cite{Enrique2002}. 



The current study presents a decision-making framework that integrates a voting system, sentiment analysis from debates, fuzzy inference systems (FIS), and consensus evaluation to perform small decision-making — determining the most suitable hotel from a set of alternatives. In our approach, experts have the flexibility to engage in GDM processes using natural language. Using sentiment analysis methods, we determine the degree to which an expert accepts a particular alternative and obtains a preference value. The methodology aims to balance quantitative evaluations with qualitative insights to enhance decision accuracy and participant satisfaction. We use FIS because of its ability to handle uncertainty and vagueness in human opinions effectively, which makes it particularly suitable because it does not rely heavily on large datasets.

The primary contribution of this paper is introducing a novel decision-making framework that integrates sentiment and emotion analysis into a fuzzy logic-based system, enhancing the effectiveness of group decision-making processes in contexts where emotional factors play a critical role.

The remainder of this paper is organized as follows: Section II reviews the relevant literature, focusing on the fuzzy approach in decision-making and sentiment and emotion analysis. Section III describes the methodology, including the preferences and voting system, sentiment analysis, and the fuzzy inference system. Section IV presents the experiment and results. Finally, Section V concludes the paper and discusses potential future research directions.




\section{Related Work}
Integrating sentiment and emotion awareness into multi-criteria fuzzy group decision-making systems is an emerging area of research. Recent papers include the SAR-MCMD method, combining multi-criteria decision-making, deep learning, and sentiment analysis \cite{2023_1}, sentiment-driven fuzzy cloud multicriteria model for online product ranking \cite{2023_2}, dual fine-tuning-based LSGDM using online reviews \cite{2023_3}, an intuitionistic fuzzy model for consensus reaching based on public opinions and expert evaluation \cite{2023_4}.

\subsection{Fuzzy Approach in Decision Making}
Group decision-making (GDM) systems use fuzzy logic \cite{Zadeh1988} and fuzzy sets \cite{Zadeh1965} to help groups make decisions \cite{Herrera2021} when there is uncertainty or ambiguity since its first appearance in the 1970 \cite{Bellman1970}. These systems allow group members to express their opinions \cite{Zadeh2002} in a way that considers the decision-making process's fuzziness and imprecision.
Fuzzy sets and logic enable the representation of imprecise or ambiguous data, which can help capture people's complex preferences and opinions in a group context. A number of works focus on applying the fuzzy approach to DM.





Kahraman et al. \cite{Kahraman2003} compared four fuzzy GDM methods for selecting a facility location: Yager's weighted goals method, Blin's fuzzy model of group decision, fuzzy analytic hierarchy process, and fuzzy synthetic evaluation. All methods aimed to choose the best facility location alternative using a multi-attribute GDM system.


GDM using incomplete fuzzy preference relations is considered in \cite{Herrera2007a}. This approach automates the consensus-reaching process without a moderator, using consensus and consistency criteria. Additionally, a feedback system provides recommendations to experts on adjusting their preferences to achieve a high degree of consensus.




\subsection{Sentiment and Emotion Analysis for Decision Making}


People usually use natural language to express their opinions during discussions \cite{Herrera1996}. Natural Language Processing (NLP) allows experts to use everyday language rather than specific formats or numbers. By extracting valuable information from large text data, NLP supports decision-making processes.

Sentiment analysis \cite{Herrera2020a}, \cite{Trillo2022}, \cite{Herrera2019a} captures the collective sentiment and emotions of group members towards a decision or option. 
Sentiment analysis techniques are widely used to understand the overall sentiment in chat conversations or social media discussions, guiding decision-making by considering the group's collective sentiment.

A study \cite{Trillo2022} proposed a large-scale GDM method to manage information from numerous experts using sentiment analysis. This approach extracts valuable insights from experts' comments, focusing on emotions related to positivity and aggressiveness. In another study \cite{Herrera2019a}, a model for GDM among experts was proposed, combining free text inputs with pairwise comparisons of alternatives. 




In dynamic contexts with numerous decision alternatives, a GDM method was introduced using a perceptual computing scheme to gather expert information \cite{Herrera2020a}. Sentiment analysis of debate texts provides valuable insights for selecting the best alternatives in each DM round. Further advancing this field, researchers developed a consensus model for dynamic social network GDM scenarios \cite{Liu2023}. Additionally, a dynamic expert weight determination approach for large-scale GDM was proposed based on social media data and sentiment analysis \cite{wan2021sentiment}. This method dynamically identifies expert weights using sentiment analysis of social media data.

Several methods were proposed to detect and analyze emotions expressed in chat conversations. This enables a deeper understanding of participants' emotional states during decision-making processes \cite{new7}, \cite{new8}, \cite{new9}, \cite{new11}.

\section{Methodology}

\subsection{Preferences, voting system}   
\begin{figure}[htp]
    \centering
    \includegraphics[width=1\textwidth]{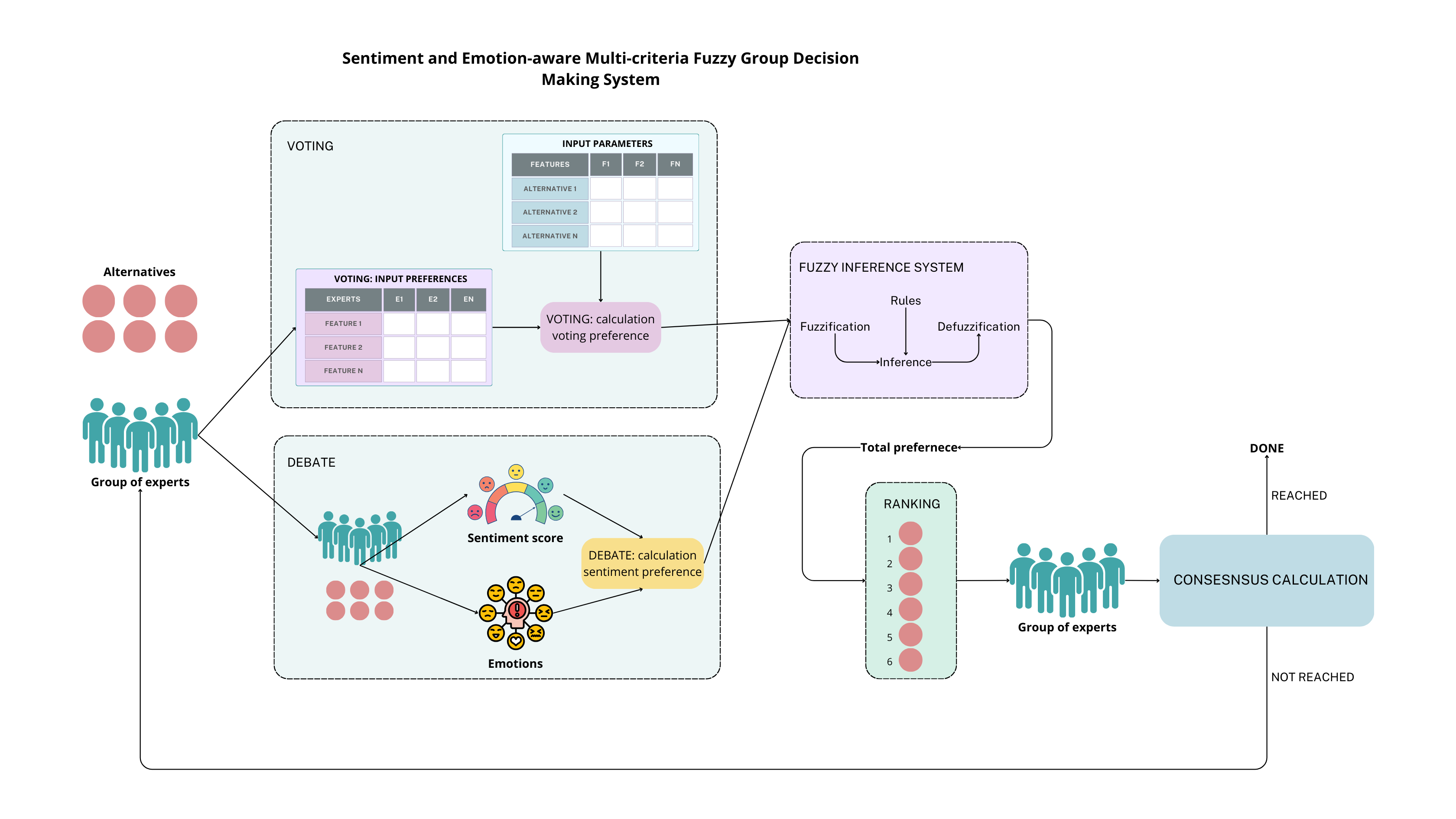}
    \caption{Sentiment and Emotion-aware Multi-criteria Fuzzy Group Decision Making System}
    \label{fig00}
\end{figure}

Figure \ref{fig00} illustrates the proposed Sentiment and Emotion-aware Multi-criteria Fuzzy GDM System system.

Let a finite set of alternatives be $X = {x_1, x_2,…, x_n}$. From the most preferred to the least preferred, these alternatives have to be ranked based on the information provided by a finite set of experts, denoted as $E = {e_1, e_2,..., e_m}$.  In addition, we have $F = {f_1, f_2, ..., f_p}$, and  $Z = {z_1, z_2,..., z_{m*p}}$, which are the features and the weights of the corresponding features, chosen by each of the experts $E = {e_1, e_2,...,e_m}$.  Because the situation in our particular setting comprises a group of friends working together to choose a hotel, the experts are considered equal. As a result of the peer-based decision-making process, every participant's opinion is given equal weight. In a GDM system, the goal is to rank the alternatives in $X$ by considering the preference values $P_k$ provided by each expert $e_j$, where $j$ ranges from 1 to $m$. In addition, we have $W = {w_1, w_2,...,  w_{n*p}}$, which are the weights of the corresponding features for each of the alternatives $X = {x_1, x_2,…, x_n}$. Preference assessments: “-1” - against, “0” - does not matter, and “1” – agreement. $Pref^{E_j}(X_i)$ – preference value of $E_j$ (expert) about $X_i$ (alternative) found using Eq. \ref{eqpref}.

\begin{equation}
Pref^{E_j}(X_i) = \sum_{i,j,k=1}^{i=n,j=m,k=p}W_k (X_i )*Z_k (E_j)
\label{eqpref}
\end{equation}

\subsection{Sentiment analysis}

Using VADER \cite{Hutto2014} from the NLTK library, Sentiment analysis provides quantitative insights from participant discussions, enhancing our understanding of their emotional responses and preferences alongside quantitative scores. We integrate emotion scores into the sentiment analysis to get the final preference score.


\begin{equation}
\text{Preference Score} = \alpha \cdot \text{Joint Sentiment Score} + \beta \cdot \text{Emotion Score}
\end{equation}
Where $\alpha$ and $\beta$ are weights that balance the influence of sentiment and emotion.

\subsection{Fuzzy Inference System}
The Fuzzy Inference System (FIS) integrates the quantitative results from the voting system with the qualitative results from sentiment analysis to compute a comprehensive total preference score for each option. 

\begin{figure*}[t!]
    \centering
    \begin{subfigure}{0.32\textwidth}
        \centering
        \includegraphics[width=\textwidth]{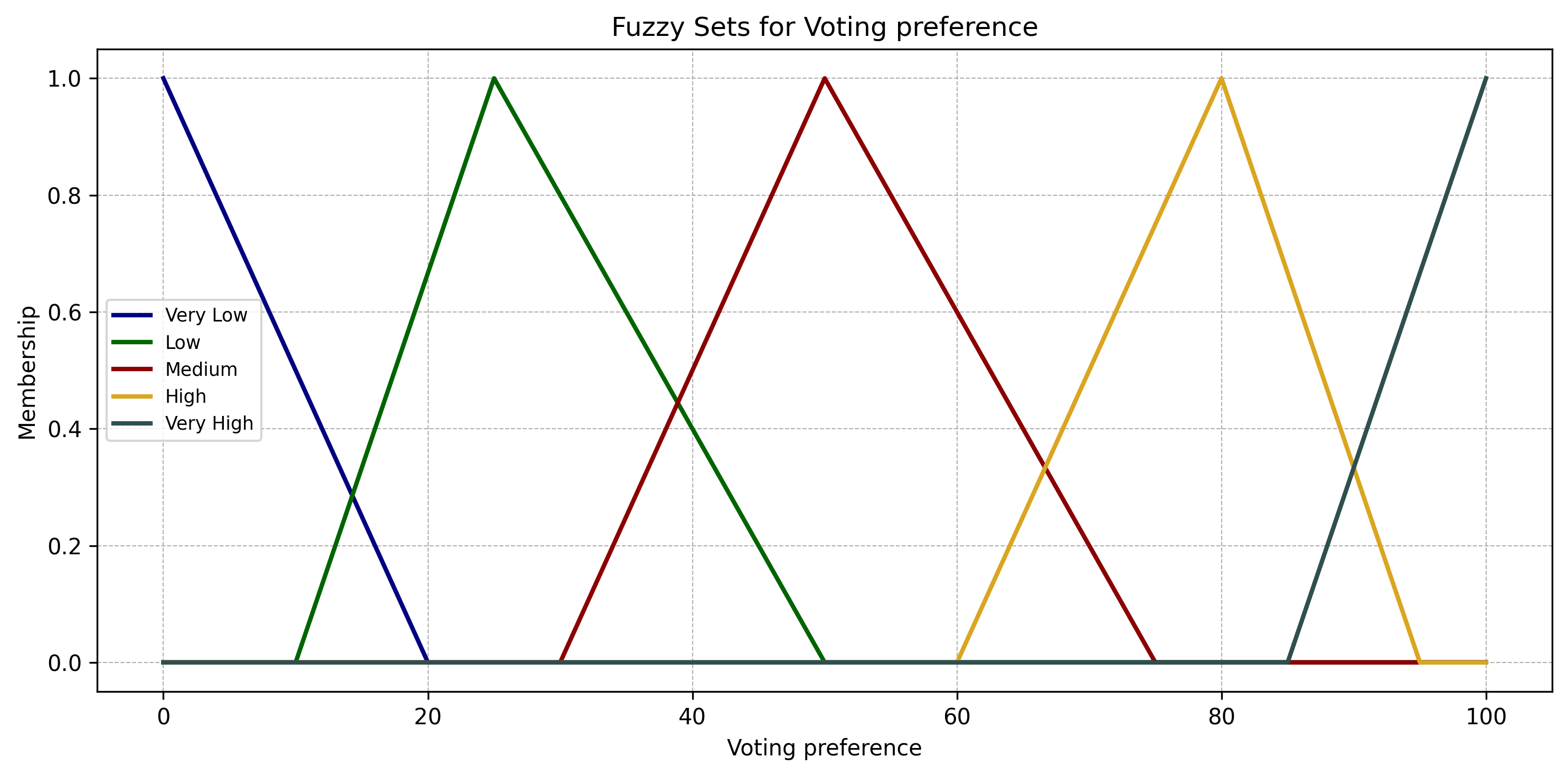}
        \caption{Voting preferences}
        \label{voting_pref_fuz}
    \end{subfigure}
    \hfill
    \begin{subfigure}{0.32\textwidth}
        \centering
        \includegraphics[width=\textwidth]{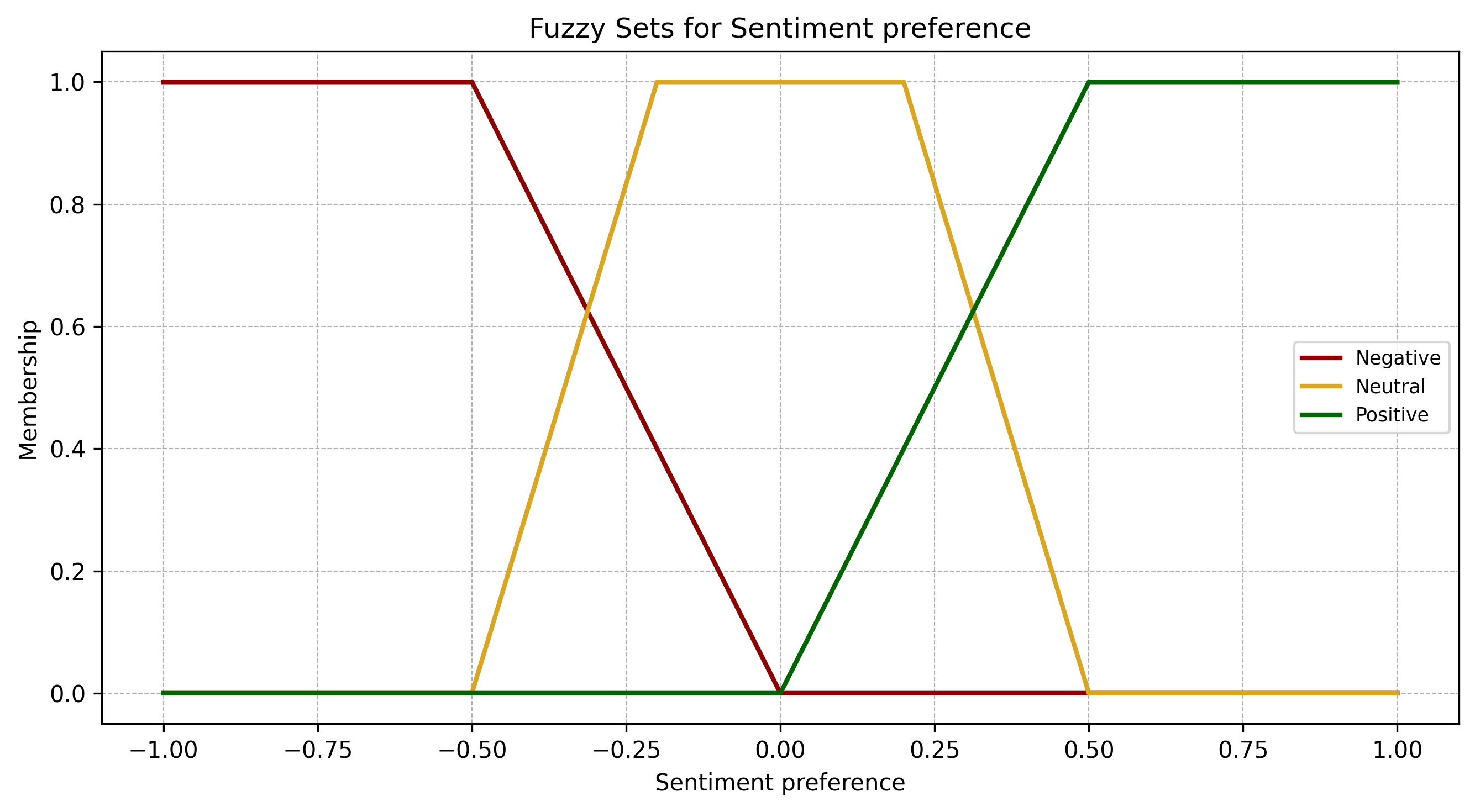}
        \caption{Sentiment preferences}
        \label{sentim_pref_fuz}
    \end{subfigure}
    \hfill
    \begin{subfigure}{0.32\textwidth}
        \centering
        \includegraphics[width=\textwidth]{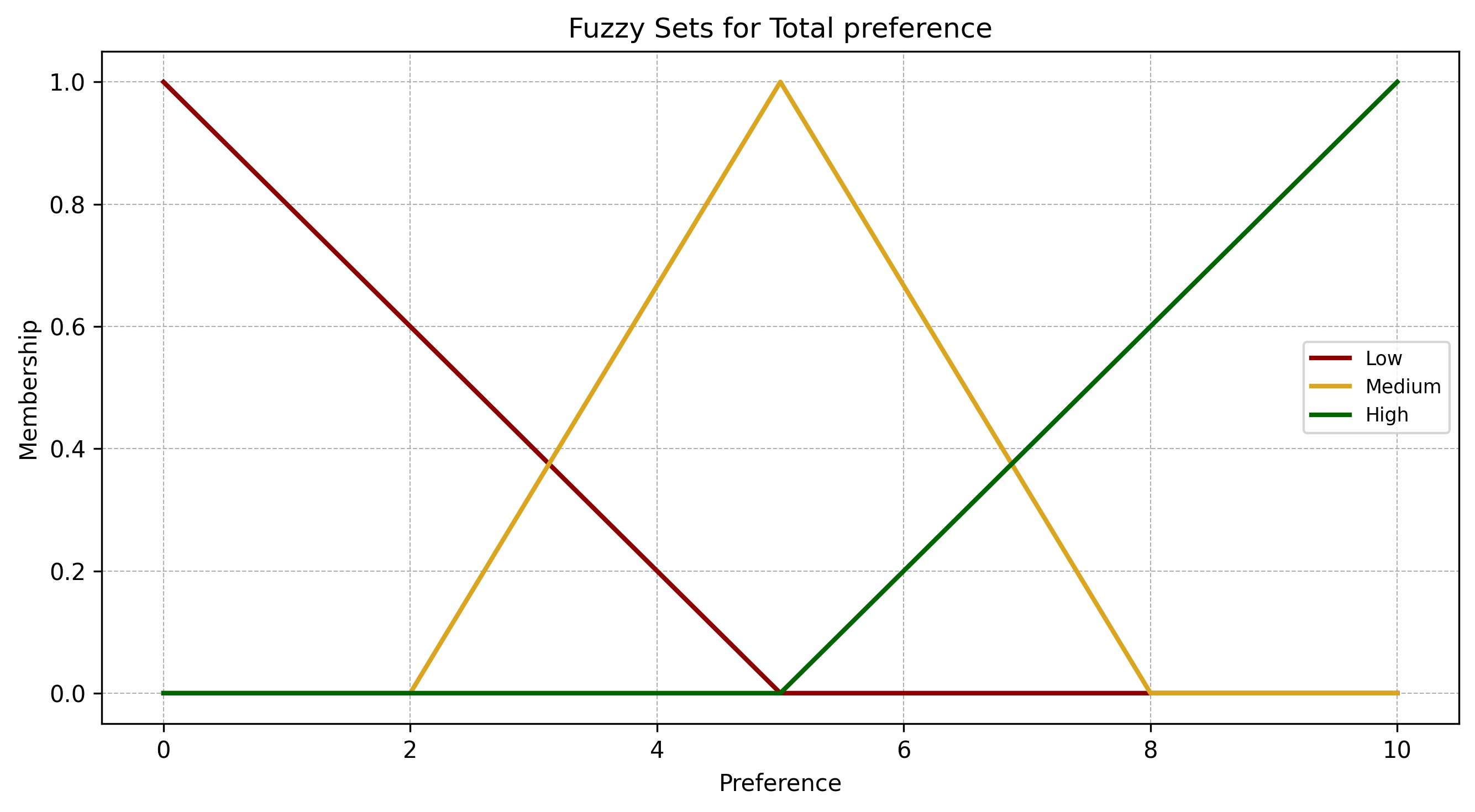}
        \caption{Total preferences}
        \label{total_pref_fuz}
    \end{subfigure}
    \caption{Input and Output Fuzzy Sets of preference scores}
    \label{fig:input_fuzzy_sets}
\end{figure*}

\begin{enumerate}
  \item \textbf{Fuzzy Input Variables}
    \begin{itemize}
      \item \textbf{Voting Preference.} This input variable represents the participants' aggregated voting scores for each alternative, scaled from 0 to 100. The membership function is shown in Figure \ref{voting_pref_fuz}.
      \item \textbf{Sentiment Preference.} This input variable represents the average sentiment score derived from the participants' qualitative discussions about each alternative, scaled from -1 to 1. The membership function is shown in Figure \ref{sentim_pref_fuz}.
    \end{itemize}
     \item \textbf{Fuzzy Output Variable.} \textbf{Total preference} represents the final preference score for each option, scaled from 0 to 10. The fuzzy sets are shown in Figure \ref{total_pref_fuz}.
\end{enumerate}






Fuzzy rules define the relationship between input and output variables using IF-THEN statements. These rules combine input levels to produce an output level, forming the fuzzy control system shown in Table \ref{tab:sentim_group_rules}.

\begin{table}[htbp]
\centering
\caption{Fuzzy rules. VL - Very Low, L - Low, M - Medium, H - High, VH - Very High.}
\begin{tabular}{|c|c|c|c|c|c|c|c|c|c|c|c|c|c|c|c|c|}
\hline
\textbf{Rule} & 1 & 2 & 3 & 4 & 5 & 6 & 7 & 8 & 9 & 10 & 11 & 12 & 13 & 14 & 15 \\ \hline
\textbf{\begin{tabular}[c]{@{}c@{}}Voting\\ Preferences\end{tabular}} & {VL} & {L} & {M} & {H} & {VH} & {VL} & {L} & {M} & {H} & {VH} & {VL} & {L} & {M} & {H} & {VH} \\ \hline
\textbf{\begin{tabular}[c]{@{}c@{}}Sentiment\\ Preferences\end{tabular}} & {L} & {L} & {L} & {L} & {L} & {M} & {M} & {M} & {M} & {M} & {H} & {H} & {H} & {H} & {H} \\ \hline
\textbf{\begin{tabular}[c]{@{}c@{}}Total \\ Preferences\end{tabular}} & {L} & {L} & {M} & {M} & {M} & {L} & {L} & {M} & {H} & {H} & {L} & {M} & {H} & {H} & {H} \\ \hline
\end{tabular}
\label{tab:sentim_group_rules}
\end{table}

\section{Experiment and Results}





We set up an experiment in which four participants make a decision about a hotel travel destination. The goal is to assess how preference analysis influences the decision-making process and outcomes, such as overall satisfaction and agreement among the participants. 
Experts will decide which alternatives are the most appropriate. Therefore, they should provide their preferences and opinions for a particular set of alternatives shown in Table \ref{tab_hotels_info}. Four participants provided their preferences for each attribute using a scale from -1 to 1, where -1 indicates a negative preference, 0 indicates indifference, and 1 indicates a positive preference (see Table \ref{tab_pref_vale_partip}). These preferences were used to establish the initial voting system, as shown in Table \ref{tab:my-table}.


\begin{table}[]
\caption{Input table: Information about hotels. 0 and 1 denote the availability, e.g., whether a hotel has a pool.}
\label{tab_hotels_info}
\begin{tabular}{|r|r|r|l|r|r|r|r|r|}
\hline
\multicolumn{1}{|c|}{\textbf{id}} &
  \multicolumn{1}{c|}{\textbf{\begin{tabular}[c]{@{}c@{}}Price, \$\\  (per week)\end{tabular}}} &
  \multicolumn{1}{c|}{\textbf{Rating}} &
  \multicolumn{1}{c|}{\textbf{Meal type}} &
  \multicolumn{1}{c|}{\textbf{\begin{tabular}[c]{@{}c@{}}Room\\ area, sq. m.\end{tabular}}} &
  \multicolumn{1}{c|}{\textbf{Star}} &
  \multicolumn{1}{c|}{\textbf{\begin{tabular}[c]{@{}c@{}}Beach  \\ access\end{tabular}}} &
  \multicolumn{1}{c|}{\textbf{\begin{tabular}[c]{@{}c@{}}  City closeness, \\ minute drive\end{tabular}}} &
  \multicolumn{1}{c|}{\textbf{\begin{tabular}[c]{@{}c@{}}Pool\end{tabular}}} \\ \hline
1 & 1165 & 7.7 & Breakfast     & 42  & 5 & 1 & 30  & 1 \\ \hline
2 & 133  & 9.6 & Breakfast     & 32  & 3 & 0 & 45  & 0 \\ \hline
3 & 783  & 8.1 & Breakfast     & 50  & 4 & 1 & 10  & 1 \\ \hline
4 & 525  & 9.2 & Breakfast     & 35  & 2 & 0 & 5   & 1 \\ \hline
5 & 4039 & 9   & All inclusive & 130 & 5 & 1 & 5   & 1 \\ \hline
6 & 186  & 6.2 & All inclusive & 42  & 3 & 0 & 3   & 0 \\ \hline
7 & 319  & 8   & All inclusive & 5   & 0 & 1 & 131 & 0 \\ \hline
\end{tabular}
\end{table}

\begin{table}[]
\caption{Input table: Preference values of participants regarding the hotel features}
\label{tab_pref_vale_partip}
\begin{tabular}{|r|r|r|r|r|r|r|r|r|}
\hline
\multicolumn{1}{|c|}{\textbf{\begin{tabular}[c]{@{}c@{}}id of \\ partic.\end{tabular}}} &
  \multicolumn{1}{c|}{\textbf{\begin{tabular}[c]{@{}c@{}}price \\ affordability\end{tabular}}} &
  \multicolumn{1}{c|}{\textbf{\begin{tabular}[c]{@{}c@{}}high \\ rating\end{tabular}}} &
  \multicolumn{1}{c|}{\textbf{\begin{tabular}[c]{@{}c@{}}avail-ty \\ of food\end{tabular}}} &
  \multicolumn{1}{c|}{\textbf{\begin{tabular}[c]{@{}c@{}}big room \\ area\end{tabular}}} &
  \multicolumn{1}{c|}{\textbf{\begin{tabular}[c]{@{}c@{}}high \\ star\end{tabular}}} &
  \multicolumn{1}{c|}{\textbf{\begin{tabular}[c]{@{}c@{}}beach \\ access\end{tabular}}} &
  \multicolumn{1}{c|}{\textbf{\begin{tabular}[c]{@{}c@{}}closenes \\ to city\end{tabular}}} &
  \multicolumn{1}{c|}{\textbf{\begin{tabular}[c]{@{}c@{}}avail-ty \\ of a pool\end{tabular}}} \\ \hline
1 & 0 & 1 & -1 & 1 & 1 & 0 & -1 & 0  \\ \hline
2 & 1 & 1 & 0  & 0 & 1 & 1 & 1  & 0  \\ \hline
3 & 1 & 1 & -1 & 1 & 1 & 0 & 0  & -1 \\ \hline
4 & 1 & 1 & 0  & 0 & 0 & 1 & 1  & 0  \\ \hline
\end{tabular}
\end{table}


The data comprises two primary components: participant preferences and hotel characteristics. We considered seven hotels, each with the following eight attributes:
\begin{itemize}
    \item \textbf{Price per week (\$):} The cost for a week's stay.
    \item \textbf{Rating:} The average user rating on a scale of 1 to 10.
    \item \textbf{Meal Type:}  Whether the hotel offers breakfast only or all-inclusive meal plans.
    \item \textbf{Room Area (square meters):} The size of the hotel room.
    \item \textbf{Star Rating:} The hotel's star classification.
    \item \textbf{Beach Access:} Binary indicator of whether the hotel has direct access to a beach.
    \item \textbf{Closeness to City (minutes):} The duration of the car ride to the city center.
    \item  \textbf{Availability of a Swimming Pool:} The hotel's swimming pool's binary indicator.
\end{itemize}



The preference score for each participant regarding each hotel is calculated by comparing the hotel's features to the group's average preferences. The steps are as follows. Continuous features such as price, rating, room area, and closeness to the city are normalized by comparing them to their respective mean values across all hotels by using the Formula $\text{Mean} = \frac{\sum_{i=1}^{n} x_i}{n}$, where $n$ is the total number of objects, $i$ - from 1 to $n$. To standardize the comparison of hotel attributes, we mapped the meal type attribute to numerical coefficients (0.33 for Breakfast and 1.0 for All Inclusive). For features already in binary form, values are taken without normalization. 

Next, using Eq. \ref{eqpref} we calculated the preference value for each expert to each alternative, shown in Table \ref{tab:my-table}. The preference scores, initially ranging from -8 to 8, are scaled to a more interpretable range of 0 to 100. Once all experts provide their preferences, these are aggregated into a collective preference matrix, representing the group's opinion. The group preference value is calculated by averaging the experts' preference values, as shown in Table \ref{tab_collect_pref}.

\begin{table}[]
\caption{Preference matrix of experts from voting}
\resizebox{\columnwidth}{!}{%
\begin{tabular}{|c|ccccccc|ccccccc|}
\hline
\multicolumn{1}{|l|}{}                  & \multicolumn{7}{c|}{\textbf{set 1}}                                                                                                                                                                                    & \multicolumn{7}{c|}{\textbf{set 2}}                                                                                                                                                                                    \\ \hline
\multicolumn{1}{|l|}{\textbf{hotel ID}} & \multicolumn{1}{c|}{\textbf{1}} & \multicolumn{1}{c|}{\textbf{2}} & \multicolumn{1}{c|}{\textbf{3}} & \multicolumn{1}{c|}{\textbf{4}} & \multicolumn{1}{c|}{\textbf{5}} & \multicolumn{1}{c|}{\textbf{6}} & \textbf{7} & \multicolumn{1}{c|}{\textbf{1}} & \multicolumn{1}{c|}{\textbf{2}} & \multicolumn{1}{c|}{\textbf{3}} & \multicolumn{1}{c|}{\textbf{4}} & \multicolumn{1}{c|}{\textbf{5}} & \multicolumn{1}{c|}{\textbf{6}} & \textbf{7} \\ \hline
\textbf{parp1}                          & \multicolumn{1}{c|}{0}          & \multicolumn{1}{c|}{1}          & \multicolumn{1}{c|}{1}          & \multicolumn{1}{c|}{0}          & \multicolumn{1}{c|}{1}          & \multicolumn{1}{c|}{-2}         & -1         & \multicolumn{1}{c|}{50}         & \multicolumn{1}{c|}{56,25}      & \multicolumn{1}{c|}{56,25}      & \multicolumn{1}{c|}{50}         & \multicolumn{1}{c|}{56,25}      & \multicolumn{1}{c|}{37,5}       & 43,75      \\ \hline
\textbf{parp2}                          & \multicolumn{1}{c|}{3}          & \multicolumn{1}{c|}{2}          & \multicolumn{1}{c|}{4}          & \multicolumn{1}{c|}{4}          & \multicolumn{1}{c|}{4}          & \multicolumn{1}{c|}{2}          & 2          & \multicolumn{1}{c|}{68,75}      & \multicolumn{1}{c|}{62,5}       & \multicolumn{1}{c|}{75}         & \multicolumn{1}{c|}{75}         & \multicolumn{1}{c|}{75}         & \multicolumn{1}{c|}{62,5}       & 62,5       \\ \hline
\textbf{parp3}                          & \multicolumn{1}{c|}{0}          & \multicolumn{1}{c|}{2}          & \multicolumn{1}{c|}{2}          & \multicolumn{1}{c|}{1}          & \multicolumn{1}{c|}{1}          & \multicolumn{1}{c|}{0}          & 0          & \multicolumn{1}{c|}{50}         & \multicolumn{1}{c|}{62,5}       & \multicolumn{1}{c|}{62,5}       & \multicolumn{1}{c|}{56,25}      & \multicolumn{1}{c|}{56,25}      & \multicolumn{1}{c|}{50}         & 50         \\ \hline
\textbf{parp4}                          & \multicolumn{1}{c|}{2}          & \multicolumn{1}{c|}{2}          & \multicolumn{1}{c|}{3}          & \multicolumn{1}{c|}{4}          & \multicolumn{1}{c|}{3}          & \multicolumn{1}{c|}{2}          & 2          & \multicolumn{1}{c|}{62,5}       & \multicolumn{1}{c|}{62,5}       & \multicolumn{1}{c|}{68,75}      & \multicolumn{1}{c|}{75}         & \multicolumn{1}{c|}{68,75}      & \multicolumn{1}{c|}{62,5}       & 62,5       \\ \hline
\end{tabular}%
}

\label{tab:my-table}
\end{table}

\begin{table}[]
\caption{Collective preference matrix from voting}
\label{tab_collect_pref}
\begin{tabular}{|c|l|l|l|l|l|l|l|}
\hline
\textbf{Hotel}                                                                 & hotel 1 & hotel 2 & hotel 3 & hotel 4 & hotel 5 & hotel 6 & hotel 7 \\ \hline
\textbf{\begin{tabular}[c]{@{}c@{}}Average Preference Score\end{tabular}} & 57,81   & 60,94   & 65,62   & 64,06   & 64,06   & 53,12   & 54,69   \\ \hline
\end{tabular}
\end{table}




Participants engaged in a debate about the pros and cons of each hotel through a chat messenger. This informal yet rich discussion provided valuable qualitative data. We utilized text analysis tools to extract sentiment and emotional scores from the chat. The sentiment score for each statement was calculated using the VADER sentiment analysis tool, which provides a compound score ranging from -1 (negative) to 1 (positive), as shown in Table \ref{tab_sentim_pref}. 
After calculating the sentiment scores for each comment, the next step is to aggregate these scores into an average sentiment score for each hotel (see Table \ref{tab_sentim_pref}).

\begin{table}[]
\caption{Sentiment preference matrix}
\label{tab_sentim_pref}
\begin{tabular}{|c|l|l|l|l|l|}
\hline
\multicolumn{1}{|l|}{} &
  \multicolumn{1}{c|}{\textbf{parp1}} &
  \multicolumn{1}{c|}{\textbf{parp2}} &
  \multicolumn{1}{c|}{\textbf{parp3}} &
  \multicolumn{1}{c|}{\textbf{parp4}} &
  \multicolumn{1}{c|}{\textbf{\begin{tabular}[c]{@{}c@{}}avg. sentiment score\end{tabular}}} \\ \hline
\textbf{hotel1} & -0,5615 & 0,9301  & -0,3561 & 0,2574  & 0,067475 \\ \hline
\textbf{hotel2} & -0,5267 & -0,1027 & 0,3612  & -0,0644 & -0,08315 \\ \hline
\textbf{hotel3} & 0       & 0,6597  & 0,8658  & 0,8765  & 0,6005   \\ \hline
\textbf{hotel4} & 0       & 0,2144  & 0,5478  & 0,7579  & 0,380025 \\ \hline
\textbf{hotel5} & 0       & 0,685   & 0,4404  & 0,2144  & 0,33495  \\ \hline
\textbf{hotel6} & -0,5267 & -0,831  & -0,3919 & -0,4497 & -0,54983 \\ \hline
\textbf{hotel7} & -0,296  & 0       & 0       & 0       & -0,074   \\ \hline
\end{tabular}
\end{table}

Emotional tones (\textit{happy, angry, surprise, sad, and fear}) were detected using the Text2Emotion library \cite{t2e}, which assigns scores to different emotions, as shown in Table \ref{tab_emoti_sentim}.
For each hotel, we calculated the average combined sentiment score. The emotion score was derived by considering the positive and negative emotions separately. It was calculated by taking the maximum of the following feelings: negative emotions included "angry," "sad," and "fear," while positive emotions included "happy" and "surprise." The difference between the positive and negative emotion scores was used to create the composite emotion score. This approach ensures that the most intense positive and negative emotions are considered, reflecting the overall emotional tone of the discussion. These Emotion and Sentiment scores were weighted and combined to form the total sentiment score: $Total sentiment Score =   Emotion Score * 0.4 + Sentiment Score * 0.6$. This weighting scheme gives more importance to sentiment analysis while still considering emotional intensity.

These weights were chosen to give slightly more emphasis to sentiment analysis due to its broader contextual relevance. This weighting scheme ensures that while sentiment analysis plays a dominant role, the emotional intensity is still substantially incorporated. In general, they can be adjusted based on the context or derived from user preferences.



\begin{table}[]
\caption{Emotion preference scores}
\label{tab_emoti_sentim}
\begin{tabular}{|l|l|l|l|l|l|l|l|l|l|}
\hline
\multicolumn{1}{|c|}{\textbf{Part.}} &
  \multicolumn{1}{c|}{\textbf{Hotel}} &
  \multicolumn{1}{c|}{\textbf{Happy}} &
  \multicolumn{1}{c|}{\textbf{Angry}} &
  \multicolumn{1}{c|}{\textbf{Surprise}} &
  \multicolumn{1}{c|}{\textbf{Sad}} &
  \multicolumn{1}{c|}{\textbf{Fear}} &
  \multicolumn{1}{c|}{\textbf{\begin{tabular}[c]{@{}c@{}}Positive \\ Emotion\end{tabular}}} &
  \multicolumn{1}{c|}{\textbf{\begin{tabular}[c]{@{}c@{}}Negative \\ Emotion\end{tabular}}} &
  \multicolumn{1}{c|}{\textbf{\begin{tabular}[c]{@{}c@{}}Emotion \\ score\end{tabular}}} \\ \hline
parp1 & hotel1 & 0    & 0,17 & 0    & 0    & 0,83 & 0    & 0,83 & -0,83 \\ \hline
parp2 & hotel1 & 0,43 & 0,14 & 0,14 & 0,14 & 0,14 & 0,43 & 0,14 & 0,29  \\ \hline
parp3 & hotel1 & 0    & 0    & 0    & 0    & 1    & 0    & 1    & -1    \\ \hline
parp4 & hotel1 & 0,4  & 0    & 0,2  & 0,2  & 0,2  & 0,4  & 0,2  & 0,2   \\ \hline
...   & ...    & ...  & ...  & ...  & ...  & ...  & ...  & ...  & ...   \\ \hline
parp4 & hotel7 & 0    & 0    & 0    & 0    & 1    & 0    & 1    & -1    \\ \hline
\end{tabular}
\end{table}

\begin{table}[]
\caption{Combine Sentiment and Emotion Scores into a final preference score and Total preference score after calculation by Fuzzy Inference System}
\label{tab_fianl_emot_senti_score}
\begin{tabular}{|l|l|l|l|l|l|}
\hline
\multicolumn{1}{|c|}{\textbf{Hotel}} &
  \multicolumn{1}{c|}{\textbf{\begin{tabular}[c]{@{}c@{}}Average Combined \\ Emotion Score\end{tabular}}} &
  \multicolumn{1}{c|}{\textbf{\begin{tabular}[c]{@{}c@{}}Average Combined \\ Sentiment Score\end{tabular}}} &
  \multicolumn{1}{c|}{\textbf{\begin{tabular}[c]{@{}c@{}}Total Sentiment\\ Score\end{tabular}}} &
  \multicolumn{1}{c|}{\textbf{\begin{tabular}[c]{@{}c@{}}Average Preference\\ Score from Voting\end{tabular}}} &
  \multicolumn{1}{c|}{\textbf{\begin{tabular}[c]{@{}c@{}}Fuzzy\\ score\end{tabular}}} 
  \\ \hline
hotel1 & -0,34 & 0,07  & -0,09 & 57,81 & 5,00 \\ \hline
hotel2 & 0,13  & -0,08 & 0,00  & 60,94 & 5,17 \\ \hline
hotel3 & 0,31  & 0,60  & 0,49  & 65,62 & 7,52 \\ \hline
hotel4 & -0,40 & 0,38  & 0,07  & 64,06 & 5,74 \\ \hline
hotel5 & -0,19 & 0,33  & 0,13  & 64,06 & 5,88 \\ \hline
hotel6 & -0,38 & -0,55 & -0,48 & 53,12 & 5,00 \\ \hline
hotel7 & -0,63 & -0,07 & -0,29 & 54,69 & 5,00 \\ \hline
\end{tabular}
\end{table}

\begin{figure*}[t!]
    \centering
    \begin{subfigure}{0.4\textwidth}
        \centering
        \includegraphics[width=\textwidth]{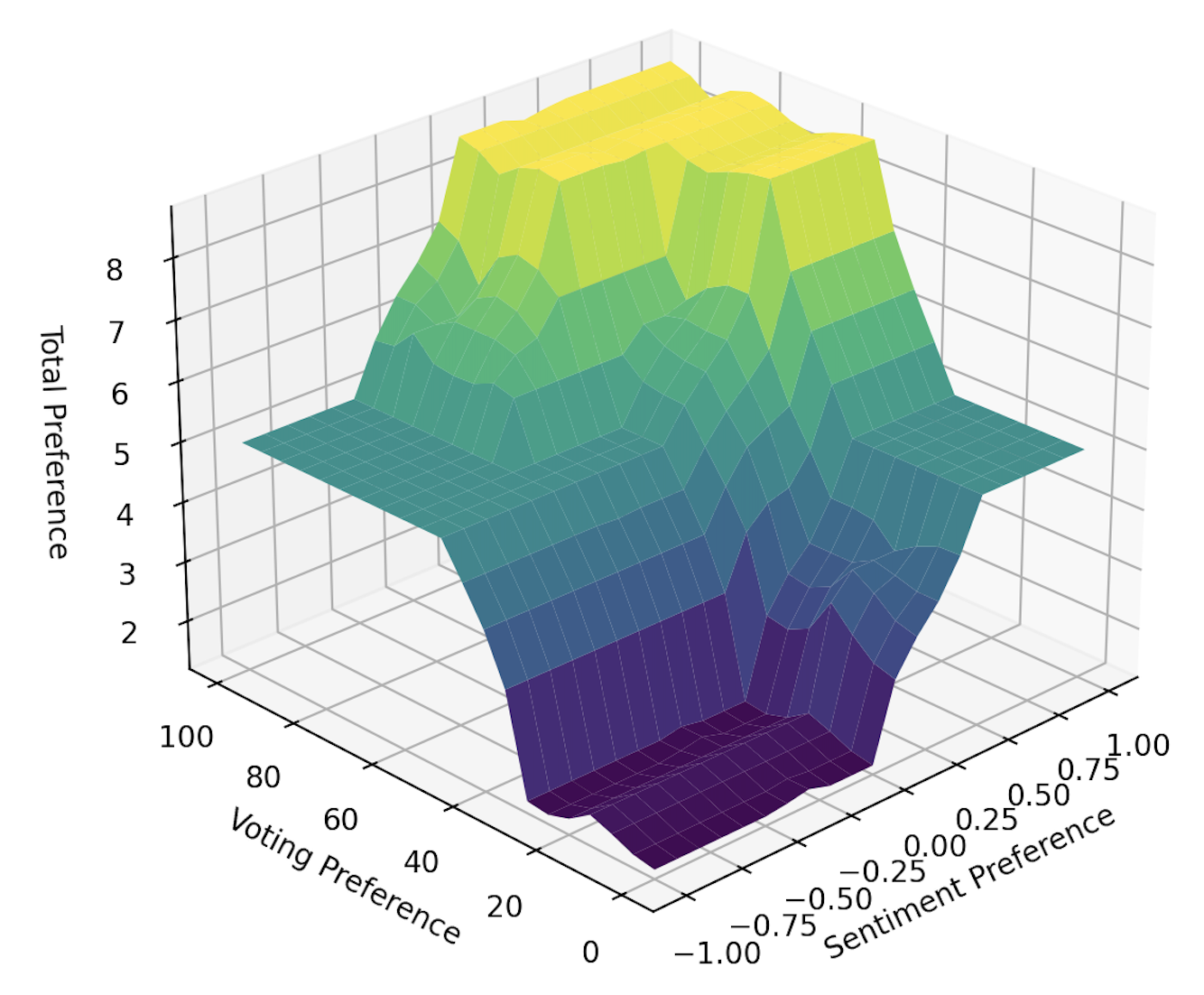}
        \caption{Preference score}
        \label{3d_graph_pref}
    \end{subfigure}
    \hfill
    \begin{subfigure}{0.4\textwidth}
        \centering
        \includegraphics[width=\textwidth]{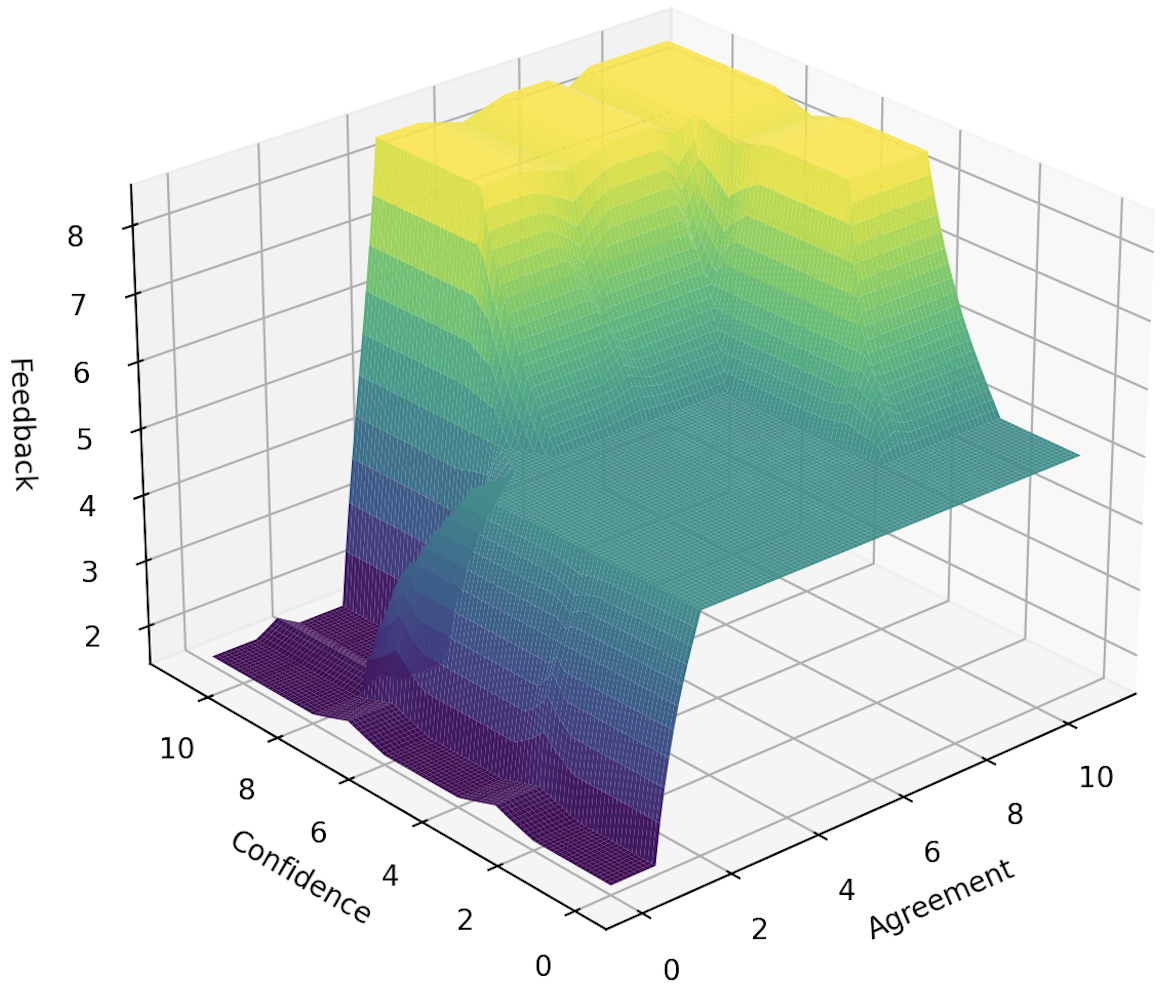}
        \caption{Feedback score}
        \label{3d_graph_feed}
    \end{subfigure}

    \caption{3D graphic representations of a set of possible fuzzy system solutions}
    \label{fig:3d_graph}
\end{figure*}

\begin{figure}[t!]
    \centering
    \begin{minipage}[t!] {\dimexpr.25\textwidth-1em}
    \centering
    \begin{subfigure}{\textwidth}
        \centering
        \includegraphics[width=\textwidth]{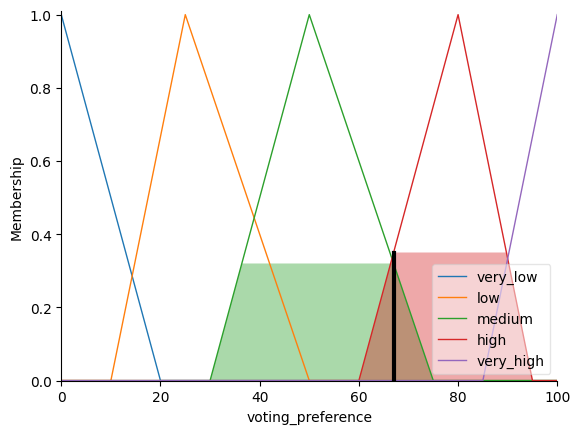}
        \caption{Voting pref-ces}
        \label{fi_voting_pref_fuz}
    \end{subfigure}
    \begin{subfigure}{\textwidth}
        \centering
        \includegraphics[width=\textwidth]{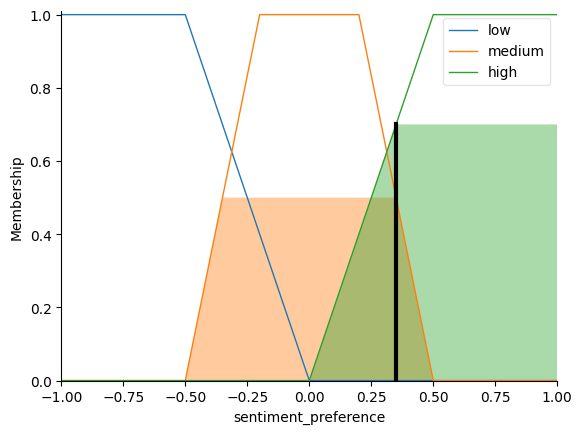}
        \caption{Sent. pref-ces}
        \label{fi_sentim_pref_fuz}
    \end{subfigure}
    \begin{subfigure}{\textwidth}
        \centering
        \includegraphics[width=\textwidth]{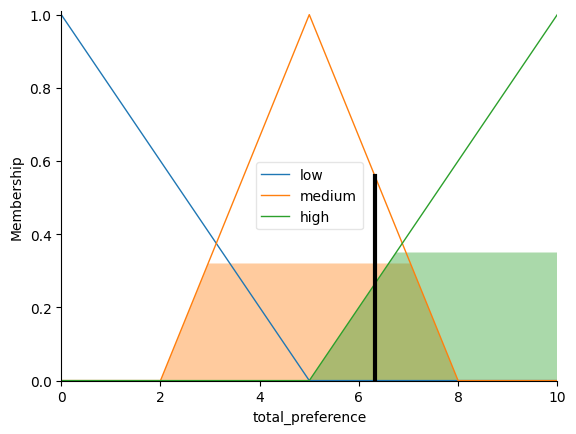}
        \caption{Total pref-ces}
        \label{fi_total_pref_fuz}
    \end{subfigure}
    \caption{Fuzzy Inference system preferences}
    \label{fig:fi_pref_fuz}
\end{minipage} 
    \hfill
    \begin{minipage}[t!] {\dimexpr.73\textwidth-1em}
        \includegraphics[width=\textwidth]{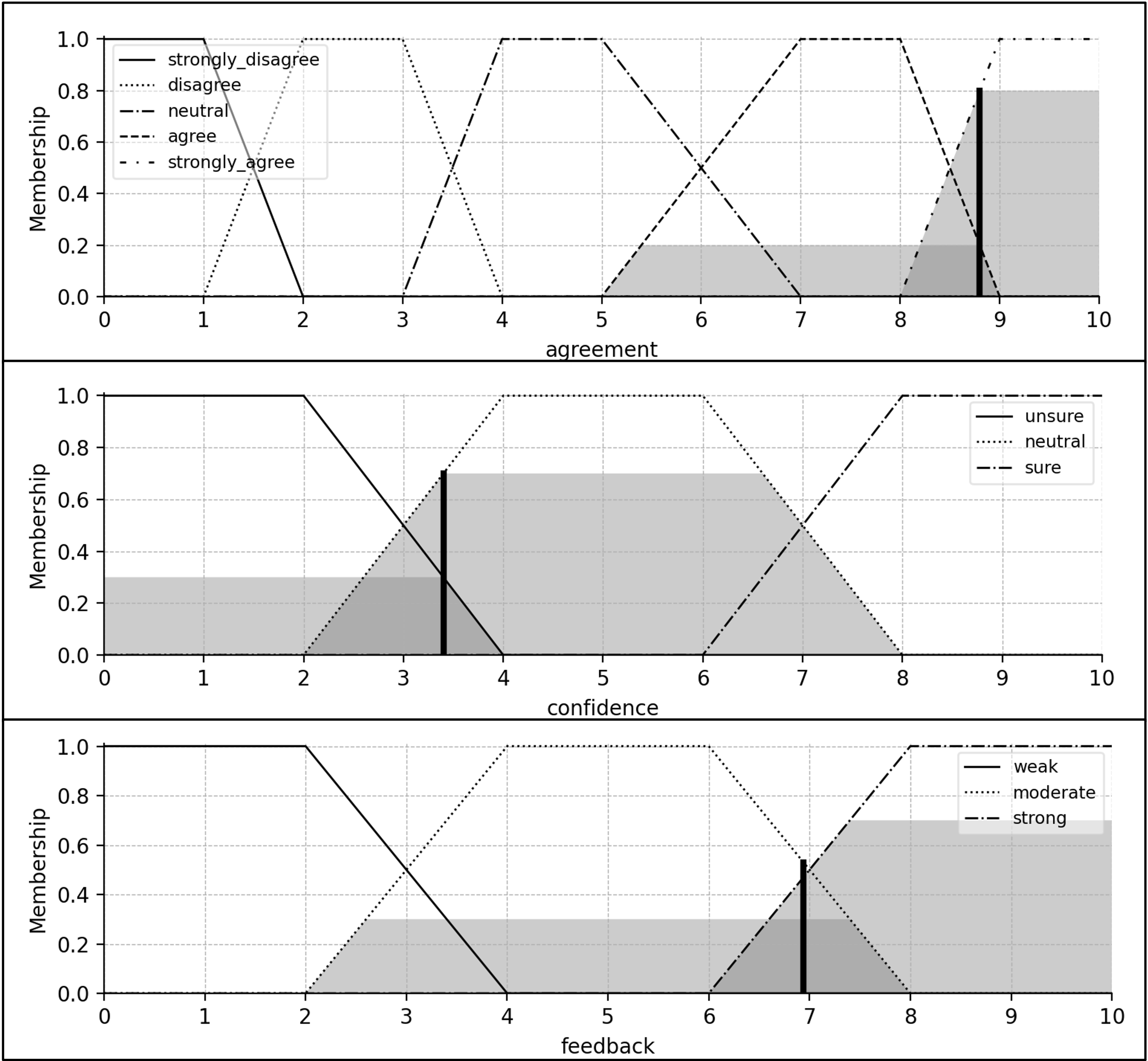}
        \caption{Visual representation of the inference process. For example, the \textit{Agreement} value is 8.8, and the \textit{Confidence} value is 3.4. As a result, we get a \textit{Feedback value} $\approx$ 6.94.}
        \label{fig11}
    \end{minipage}
    \label{fig:combined_layout}
\end{figure}

Next, we used the Fuzzy inference system to simulate and compute the total preference score (see Fig. \ref{3d_graph_pref}) based on given input values for voting preference and sentiment preference (see Fig. \ref{fig:fi_pref_fuz}), total preference values shown in Table \ref{tab_fianl_emot_senti_score}. The alternatives are ranked based on the preference levels provided by the experts.




After selecting a hotel, participants provided feedback on their agreement and confidence levels regarding the decision (see Fig. \ref{3d_graph_feed}). These values were computed using another fuzzy inference system to compute a feedback score (see Fig. \ref{fig11}). The feedback system evaluated the consistency and confidence of the participants' opinions, as shown in Table \ref{tab_feedback}. 
Our feedback system is non-monotonic, but this behavior is justified and appropriate given the interpretation of "confidence" as the confidence level in the agreement. Our rules capture the nuanced relationship between agreement, confidence in that agreement, and feedback, leading to outputs that make sense given the inputs.
Consensus metrics, including the interquartile range (IQR) and mean consensus level, were used to evaluate the overall agreement among participants, as shown in Table \ref{tab_consensus}. The consensus level was classified as \textit{high, medium}, or\textit{ low} based on the mean consensus score. This classification helped understand the group's satisfaction with the decision-making process.



\begin{table}
\begin{minipage}{0.50\textwidth}
    \centering
\caption{Participants Feedback on Hotel Recommendation}
\label{tab_feedback}
\begin{tabular}{|r|r|r|r|}
\hline
\multicolumn{1}{|c|}{\textbf{\begin{tabular}[c]{@{}c@{}}id of\\ participant\end{tabular}}} &
  \multicolumn{1}{c|}{\textbf{\begin{tabular}[c]{@{}c@{}}Agreement \\ level (0-10)\end{tabular}}} &
  \multicolumn{1}{c|}{\textbf{\begin{tabular}[c]{@{}c@{}}Confidence\\  level (0-10)\end{tabular}}} &
  \multicolumn{1}{c|}{\textbf{\begin{tabular}[c]{@{}c@{}}Feedback \\ value\end{tabular}}} \\ \hline
1 & 5  & 7  & 6.24 \\ \hline
2 & 9  & 8  & 8.44 \\ \hline
3 & 8  & 7  & 6.24 \\ \hline
4 & 10 & 10 & 8.44 \\ \hline
\end{tabular}
    
\end{minipage}
\hfill
\begin{minipage}{0.40\textwidth}
    \centering
   \caption{Consensus score}
\label{tab_consensus}
\begin{tabular}{|c|c|c|}
\hline
\textbf{\begin{tabular}[c]{@{}c@{}}IQR\end{tabular}} & \textbf{Mean} & \textbf{Consensus level} \\ \hline
2.19                                                                          & 7.34          & high                     \\ \hline
\end{tabular}
\end{minipage}

\end{table}

\section{Conclusion}
We presented a novel decision-making framework integrating voting preferences, debate analysis, and fuzzy inference systems.  The main contribution of this paper is integrating sentiment and emotion analysis into a fuzzy logic-based system, thereby making traditional group decision-making processes emotion-aware, particularly in situations where emotional factors are crucial. An example of the application of our framework is selecting a hotel.  The high consensus level among participants further validates the proposed methodology. This approach can be applied to various GDM scenarios, providing a reliable method for achieving consensus. As for future works, we aim to explore the system's applicability in more complex contexts and further refine its components to increase its generalizability. While our current work shows that integrating sentiment and emotion analysis into a multi-criteria fuzzy GDM system is feasible, as demonstrated with the hotel selection example, we plan to explore its effectiveness in more varied and complex decision-making scenarios in future research. We also plan to conduct a comprehensive sensitivity analysis and compare our approach with other decision-making frameworks to evaluate its effectiveness and reliability more rigorously. 

The proposed system integrates sentiment and emotion analysis, balancing both quantitative and qualitative inputs, making it well-suited for real-world scenarios like the hotel selection case study. However, its complexity due to multiple components can make implementation challenging. The system's effectiveness may also vary in different decision-making contexts, especially given the simplicity of the current case study. Additionally, the approach depends heavily on the quality of textual data, which could affect outcomes if the data is incomplete or biased.

In the future, we will explore integrating our framework with conventional multi-criteria decision support techniques like TOPSIS, ELECTRE, and PROMETHEE to enable a thorough comparison and assessment of its efficacy in various decision-making scenarios.

\section*{Acknowledgment}
This research has been funded by the Science Committee of the Ministry of Science and Higher Education of the Republic of Kazakhstan (Grant No. AP22786412)





\bibliographystyle{unsrt}
\bibliography{export}

\end{document}